# Do Current Video LLMs Have Strong OCR Abilities? A Preliminary Study

**Yulin Fei**[1,2*], **Yuhui Gao**[1*], **Xingyuan Xian**[2*]

**Xiaojin Zhang**[1], **Tao Wu**[1], **Wei Chen**[1†]

[1]School of Software Engineering, Huazhong University of Science and Technology
[2]Information Hub, Hongkong University of Science and Technology (Guangzhou)
**Correspondence:** lemuria_chen@hust.edu.cn

## Abstract

With the rise of multi-modal large language models, accurately extracting and understanding textual information from video content—referred to as video-based optical character recognition (Video OCR)—has become a crucial capability. This paper introduces a novel benchmark designed to evaluate the video OCR performance of multi-modal models in videos. Comprising 1,028 videos and 2,961 question-answer pairs, this benchmark proposes several key challenges through 6 distinct sub-tasks: (1) Recognition of text content itself and its basic visual attributes, (2) Semantic and Spatial Comprehension of OCR objects in videos (3) Dynamic Motion detection and Temporal Localization. We developed this benchmark using a semi-automated approach that integrates the OCR ability of image LLMs with manual refinement, balancing efficiency, cost, and data quality. Our resource aims to help advance research in video LLMs and underscores the need for improving OCR ability for video LLMs. The benchmark will be released on https://github.com/YuHuiGao/FG-Bench.git.

## 1 Introduction

The past two years have witnessed a surge in the development of visual large language models (visual LLMs), with significant progress made particularly in the domain of image-based LLMs. This area has seen the emergence of numerous models and datasets designed to enhance the understanding and generation of text from images. Simultaneously, the field of video large language models has been advancing at a comparable pace behind, with pioneering models such as Video-ChatGPT (Maaz et al., 2023) and Video-LLaVA (Lin et al., 2023) leading the way. These models have laid the foundation for integrating video content with language models, facilitating a deeper comprehension of visual and temporal information.

Alongside these developments in LLMs model structure, several datasets in the field of video understanding like MVbench (Li et al., 2024a) and Video-Instruct (Maaz et al., 2023), have been specifically crafted to meet the demands of video LLMs. With previous video QA datasets such as MSRVTT-QA (Xu et al., 2017) and TGIF-QA (Jang et al., 2019), they have fostered further innovation in this burgeoning field.

While these datasets have significantly driven innovation in video-based reasoning and question answering, an equally critical area—optical character recognition (OCR) within videos, or known as video OCR—is respectively less studied. In the field of Image OCR, there have been several benchmarks and datasets including OCR Bench (Liu et al., 2023b), MTVQA (Tang et al., 2024) and EST-VQA Dataset (Wang et al., 2020). Commercial Models like GPT-4V (Achiam et al., 2023) and Gemini 1.5 (Reid et al., 2024) have already demonstrated exceptional capabilities, outperforming on these established benchmarks. Open-source large models are also striving to improve image OCR capabilities, with some models such as TextMonkey (Liu et al., 2024) and mPlug-DocOwl1.5 (Ye et al., 2023) specifically developed for reading text in various images.

Why we need video-OCR abilities in video LLM? They enable applications far beyond traditional Image LLMs by leveraging temporal information and dynamic scene understanding. They allow precise text tracking across frames, such as identifying moving text on vehicles or dynamic signage, essential for smart transportation. In education, they extract and summarize evolving textual content like lecture slides, enriching learning resources. For security, they monitor real-time updates, such as warning messages or fluctuating counters, enhancing surveillance. In entertainment, they enable dynamic subtitle generation, context-aware ad analysis, and multi-language text translation. These abilities make Video LLMs indispensable for complex video-centric tasks across diverse domains.

*These authors contributed equally.
†Corresponding Author.

However, up to now there is still few video-OCR benchmark or dataset particularly designed for video-LLMs. In this scenarios, this paper construct the first ever known benchmark for video OCR to our knowledge, with 6 sub-tasks to evaluate models' ability on (a) Text Recognition (TR) (b) Semantic Understanding (SU) (c) Spatial Relation (SR) (d) Movement Detection (MD) (e) Text Attribute Recognition (TAR) (f) Temporal Localization (TL). The paper also provides an analysis of current visual LLMs, particularly video LLMs.

Overall, our contributions are below 1). We develop a benchmark with 6 sub-tasks across different aspects of video-OCR capabilities for testing multi-modal LLMs' OCR performance in video. This is the first big enough benchmark for evaluating visual LLMs' video OCR performance in various dimensions; 2). To construct this benchmark, we designed a semi-automatic method that combines utilizing the excellent OCR ability of image LLM and human's manual check; We designed that specifically test video LLMs ; 3). We evaluate several multi-modal LLMs on this benchmark, including both image LLMs and video LLMs. In addition, some image LLMs which only support single-image input are also being tested on this benchmark via some tricky processes and demonstrate acceptable performance.

## 2 Related Work

### 2.1 Image LLMs

Recent developments in image large language models (LLMs) have significantly advanced multimodal AI. Flamingo (Alayrac et al., 2022) by DeepMind utilizes perceptual resampling and gated cross-attention for improved visual-language integration. The LLaVA series (Liu et al., 2023a) enhances dialogue generation through visual instruction tuning. GPT-4V (Achiam et al., 2023) extends the GPT series with strong multimodal and example-based learning, while Gemini 1.5 (Reid et al., 2024) from Google excels in long-context processing and multilingual translation. These models advance practical applications in education, healthcare, and more.

### 2.2 Video LLMs

Video LLMs have made significant advancements, with several notable models enhancing multimodal capabilities. For instance, Video-LLaVA (Lin et al., 2023) effectively aligns visual and textual data, addressing challenges in integration. Similarly, Video-ChatGPT (Maaz et al., 2023) showcases advanced comprehension by leveraging CLIP ViT-L/14 (Radford et al., 2021) and Vicuna-v1.1 (Chiang et al., 2023). Chat-UniVi (Jin et al., 2024) adopts a unified approach to processing both video and text, while Video-LLaMA2 (Cheng et al., 2024) excels in narrative generation and content analysis. Notably, many of these video LLMs are built upon the foundations of image LLMs, reflecting the influence of existing visual models in their development.

### 2.3 Image OCR Benchmarks & Datasets

Key benchmarks and datasets for Optical Character Recognition (OCR) include DT-VQA (Zhang et al., 2024), focusing on Visual Question Answering, and MTVQA (Tang et al., 2024), which enriches multimedia VQA tasks. The ESTVQA (Wang et al., 2020) dataset integrates structured text and visuals, and OCRbench (Liu et al., 2023b) explores OCR in large multimodal models. These resources advance OCR research by integrating text and visual data.

### 2.4 Video OCR Datasets Pre-LLM Era

Video OCR has been enhanced by specialized datasets such as News-VideoQA (Jahagirdar et al., 2023), which provides QA pairs from news videos, and BOVText (Wu et al., 2021), offering bilingual data. RoadText 1K (Reddy et al., 2020) focuses on text detection in driving videos, while M4-ViteVQA (Zhao et al., 2022) addresses visual QA with extensive clips. These datasets push the boundaries of video text recognition and analysis.

## 3 Benchmark Tasks for Video OCR

First, we need to distinguish between video OCR and image OCR. It is essential to recognize both their similarities and differences.

#### Differences Between Image OCR and Video OCR in LLMs

**Dynamic Information and Motion-Related Tasks** Image OCR datasets typically consist of static images without dynamic information. Consequently, Image LLMs are neither required to rec-ognizing nor capable of identify the movement of OCR objects. They also cannot leverage prompts related to motion information for OCR tasks. For example, an Image LLM cannot answer a question like *"What is the text written on the billboard that gradually becomes visible as the camera pans?"* because it lacks the temporal and contextual understanding needed to process the sequence of frames where the text is incrementally revealed. Addressing the movement of OCR objects and handling motion-related tasks represent entirely new challenges for Video OCR within LLMs.

**Temporal Information and Localization Tasks** Images in Image OCR datasets generally lack time dimension, meaning Image LLMs cannot extract or reason about the temporal characteristics of OCR objects. Tasks such as locating the timestamps when OCR objects appear or disappear, or measuring their duration on screen, are outside the capabilities of Image LLMs. Handling temporal information and tackling time-sensitive tasks, such as temporal localization of OCR objects, are groundbreaking challenges for Video OCR in LLMs.

**Integrating Multi-Frame Information** Video LLMs are designed to extract and integrate visual features across multiple frames, allowing for more robust recognition of OCR objects despite temporal disruptions. Image OCR datasets typically involve single-frame inputs, so Image LLMs do not and cannot utilize information from multiple frames to improve OCR object recognition. For example, When text in a video is intermittently obscured and reappears across different frames, the ability to correlate the text across frames and recognize it as the same object is crucial. However, this capability has not been explored in Image LLMs and remains an open research challenge in the context of Video LLMs.

These distinctions highlight the unique challenges and opportunities in the development of Video OCR capabilities within Video LLMs, extending far beyond the scope of traditional Image LLMs.

### 3.1 Tasks Introduction

To evaluate the ability of visual LLMs in the field of video OCR, we develop this benchmark which consist 1028 videos and 2961 question-answer pairs, divided into 6 sub-tasks. Here is the figure (fig 1) of question-answer pairs example for the 6 sub-tasks:

**Text Recognition** This task involves identifying text from video frames, accurately recognizing characters, words, phrases, and sentences that appear within the video. It is derived from Image OCR tasks but differs significantly due to the dynamic nature and temporal information of video. As mentioned in last subsection, video OCR tasks must consider continuous frames over time. This introduces the challenge of leveraging temporal information and dynamic changes, as well as the need to integrate information across multiple frames. For instance, in a video showing a moving vehicle with a license plate, a text recognition task might involve identifying the license plate number as it moves across the screen, and text may be partially obscured in certain frames. Another example is a prompt like *"What is the text displayed on the digital counter in a bank that changes over time?"*. Here, the model must track the changing text across multiple frames, recognizing the sequence of numbers or words and understanding the temporal progression to maintain an accurate representation of the text.

**Semantic Understanding** This task involves comprehending the meaning and context of detected text in videos. Visual LLMs rely on semantic information for text recognition in OCR tasks (Liu et al., 2023b), and for effective OCR performance in video, a model must not only recognize the text but also understand its meaning within the context. For example, in a video showing a supermarket scene with merely a price displayed on a billboard, a prompt like *"What is the price of the apple?"* requires the model to not only read the text on the billboard but also understand its relevance to the video's context, such as identifying the fruit related to the price. This task goes beyond simple text recognition and requires the model to integrate both visual and textual information to accurately interpret the meaning.

**Spatial Relation** This task examines the spatial relationships between text elements and other objects in the video frame, focusing on their relative positions, alignments, and interactions. Visual LLMs need to recognize these relationships. For example, the most common question would be *"Where is the text {Given_Text} located?"*, models sometimes should response with a certain object with the text. While multiple frames provide richer information for better judgment, they also increase the complexity due to changing visuals.

**Text Attribute Recognition** This task involves identifying Text visual Attributes including script,

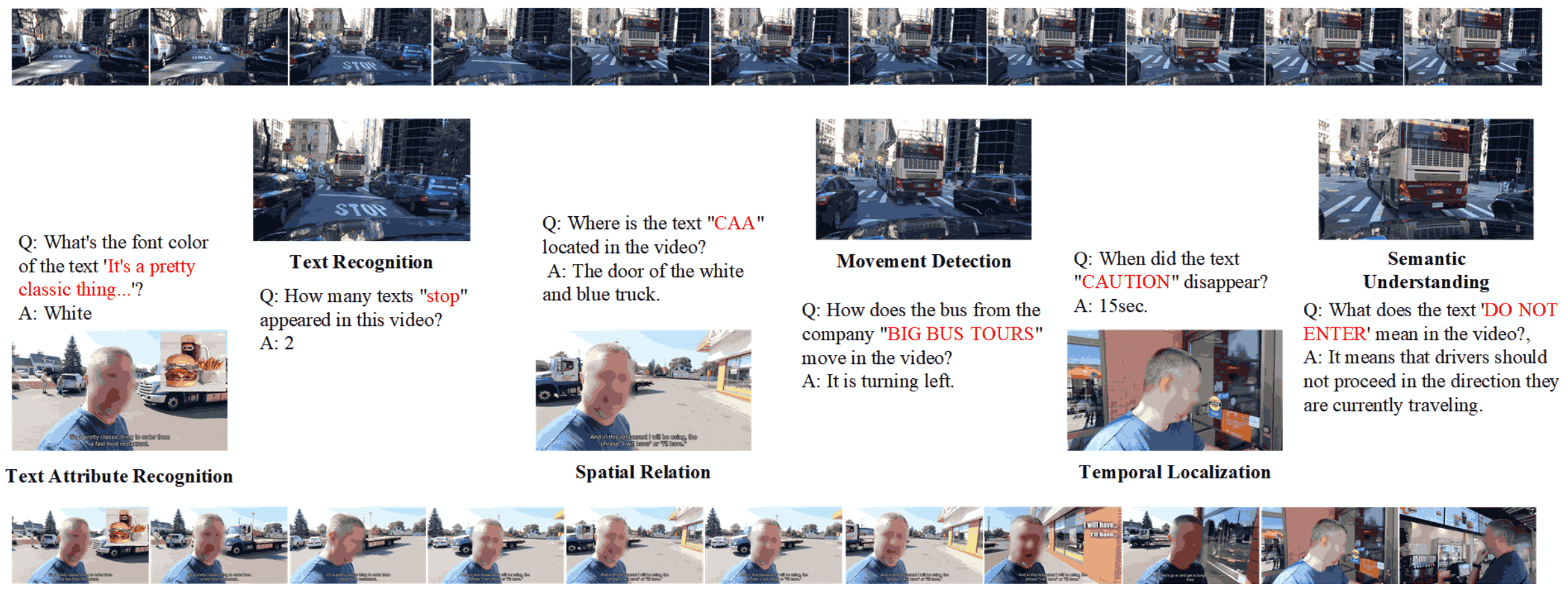

Figure 1: Sample question-answer pairs in 6 sub-tasks. Red words are target text.

language and color of the text appearing in the video, recognizing different writing systems, and possibly handling multilingual content. Among these tasks, the recognition of colors and fonts takes up the majority, while the recognition of language types is less emphasized, as identifying language types is strongly tied to the outcomes of Text Recognition and Semantic Understanding.

**Movement Detection** This task focuses on detecting and tracking text movement in videos, known as *Text Tracking*, a specialized aspect of video OCR. Unlike traditional methods that rely on bounding boxes and MOTA metrics, our approach describes text movement using natural language, as current visual LLMs typically do not output bounding boxes or coordinates directly. For example, in a video of a car with a visible direction sign, a prompt like *"Which direction is the car with license plate 'T689273' heading?"* would require the model to locate both the text "T689273" and "West, Kennedy Airport," extract visual features of the car and its movement, and then respond with *"The car is head- ing west towards Kennedy Airport"*. We convert these descriptions into a multiple-choice format for easier evaluation.

**Temporal Localization** This task focuses on identifying and localizing when specific text appears, disappears, and how long it remains visible in a video. We anticipate that current video LLMs may struggle with this task due to their architectures, which prioritize capturing temporal sequences of objects, actions, or events over direct timeline tracking. Additionally, the limited number of frames sampled by many models could further hinder accurate timestamp localization. We format this sub-task as multiple-choice, with most options as 2-second intervals or right-unbounded time ranges.

## 4 Benchmark Construction

As inspired, we first decided to produce question-answer pairs based on the existed traditional video-OCR datasets. However, the form of those datasets are conventional, not in natural language. Every OCR object is stored in part of json files, containing text content, corresponding text ID, coordinates of bounding box in frames. See figure 8.

### 4.1 Semi-Automatic Process

To manufacturing data efficiently and in high quality, we developed a semi-automatic procedure with 3 stages that leverage powerful OCR capability of certain image LLM. We will use a pseudo-code and a diagram in appendix to illustrate these 3 stages clearly. See the pseudo-code 1 and figure 9:

**Stage 1** First, we select only 4 frames evenly as backup for each OCR object in video. This is because selecting 4 frames strikes a balance between computational efficiency and effectiveness. Choosing more frames would significantly increase the inference time of image LLMs. While for many OCR objects, their presence in the video is brief (several seconds usually), and 4 frames are sufficient to capture necessary information about these objects. Then, InternVL 1.5 (Chen et al., 2024b) is deployed to summarize all visual information and text objects in 4 frames to get **4 OCR Contextual Captions**. A OCR Contextual Caption is a description supposed to involve all OCR objects and as much as possible other visual elements in the frame.

This process mainly aims to convert as much visual information to natural language information as possible due to the situation that multi-modal LLMs seems still dealing with text better than other modality. Besides, it should be noticed that for a single image, all relevant information about the OCR objects present, ranging from texts' basic visual details to their locations within the image, are input into the LLM. This can induce visual LLM to successfully uncover and express the relationships between interrelated OCR objects in its responses.

**Algorithm 1** Video-OCR Prompt Process (See detailed diagram 9 in Appendix)

**Input**: Video frames $\{F_i\}_n$ with traditional Video-OCR dataset $D$
**USED MODEL**: InternVL-1.5, GPT-3.5
**for** each OCR object $d$ in $D$ **do**
  **Stage 1:Generate OCR Contextual Caption for each frame**
  **for** each frame $F_i$ **do**
    Extract all other OCR objects and their positions. Construct *Prompt 1*
    *OCR Contextual Caption* = InternVL-1.5(*Prompt 1* + frame $F_i$)
  **end for**
  **Stage 2: Generate 4 OCR Detailed Captions for OCR object** $d$
  **for** each frame $F_i$ **do**
    *OCR Detailed Caption* for $d$ = InternVL-1.5(*OCR Contextual Caption* + *Prompt 2*)
  **end for**
  **Stage 3: Aggregate OCR Detailed Captions**
  1 combined *OCR Detailed Caption* = GPT-3.5(*Prompt 3* + 4 *OCR Detailed captions*)
  **Stage 4: Question-Answer Pairs**
  6 qa pairs = GPT-3.5(*Prompt 4* + combined caption + few shot cases)
**end for**

**Stage 2** Each Generated OCR Contextual Caption in last stage is put into InternVL 1.5 with corresponding frame and a delicately designed prompt to get **1 OCR Detailed Captions for each OCR object**. A OCR Detailed Caption is specific description of certain OCR object containing its text content, semantic meaning, movement information, timestamps about when it appear and vanish, visual attribute like color and font. Notably again that this process is conducted once for each frame. So there will be 4 OCR Detailed Captions that are generated.

**Stage 3** After getting 4 OCR Detailed Captions for the 4 sampled frames, we follow the principle of the majority and use GPT-3.5-Turbo to synthesize different captions into 1 single caption, which contains thorough information of the OCR object, ranging from its semantic content to spatial-temporal information. Then the synthesized caption will be given GPT-3.5-Turbo to generate 6 question-answer pairs in diverse sub-tasks.

### 4.2 Human Annotation & Refinement

Expert human annotators check the correctness of manufactured answer and question's logic of the question-answer pairs. Then revise those questions and answers strictly following the rules below:

**Task's Type-Question Alignment** For each question-answer pair, Human annotators first need to check whether the question is matched with type of sub-task. If not, correct the question-answer pair to match the sub-task type.

**Answerability** For questions in dataset, it is necessary to ensure every question is answerable, i.e. there must be 1 and only 1 correct answer. To achieve this goal, human need to clarify the descriptions and references of ocr objects.

**Factual Correctness** Human annotators need to revise the factual error which exist in questions and answers.

**Visual Dependence** Human annotators have to make sure the questions can only be answered after watching corresponding video.

**Simplicity** Human annotators must simplify ground truth answers across sub-tasks like Text Recognition, Attribute Recognition, Spatial Relation, and Temporal Localization. For Text Recognition and Attribute Recognition, answers should be in their simplest form, without prefixes or introductions. In Movement Detection and Spatial Relation, objects should be described by their most visually prominent features. For Temporal Localization, answers should be reduced to numbers and time units only.

**Quality Review** Another data annotator who did not participate in the data cleaning process will review the cleaned data one by one according to the above standards to verify the quality of the cleaned data and ensure that the data has been properly cleaned. Questions that do not meet those criterions are returned to the original annotators for revision.

### 4.3 Convert to Multiple Choices

In the Temporal Localization and Movement Detection sub-tasks, to lower task difficulty, inspired by Video-MME (Fu et al., 2024), where ground truth answers are provided as multiple-choice options (A, B, C, and D), we adopted a similar approach. This format not only improves model accuracy but also simplifies evaluation. For Temporal Localization, each option represents a time interval rather than a specific timestamp, with intervals set at 2 units apart. In Movement Detection, by giving the question and correct answer, GPT-3.5 generates three distractor options.

### 4.4 Analysis

After these above processes data is succeed to be manufactured. To get more details of data, please see many statistics results in appendix A.

This semi-automated pipeline is inspired by the Video-ChatGPT (Maaz et al., 2023) approach, which utilizes Katna for keyframe extraction, Tag2Text for tagging, and a combination of BLIP-2, GRiT, and GPT-3.5 for captioning. In contrast, our method simplifies this by using only one open-sourced visual LLM with sufficient OCR capabilities and GPT-3.5, achieving a favorable balance between cost and efficiency. This efficiency is due to: (a) InternVL 1.5 being open-source, reducing API costs compared to GPT-4V, and requiring only a single A6000 GPU for deployment; and (b) InternVL 1.5's strong performance in OCR Bench (Liu et al., 2023b) and captioning tasks, despite having fewer parameters than GPT-4V or Gemini 1.5.

## 5 Experiment

In this section, we delineate the process of evaluating several visual LLMs.

### 5.1 Evaluation Procedure

#### 5.1.1 What models are evaluated?

We evaluate the performance of various multi-modal large language models on the VideoOCR Benchmark, including video LLMs such as Mini-GPT-4-video (Ataallah et al., 2024), VILA-1.5-3B-video (Lin et al., 2024), Video-LLama2 (7B and 72B) (Cheng et al., 2024), Chat-UniVi-V1.5 (Jin et al., 2024), Video-CCAM (**4B** and **14B**) (Team, 2024), MiniCPM-V 2.6 (Yao et al., 2024), ShareGPT4Video (Chen et al., 2024a), and Qwen2-VL-7B (team, 2024), most of which have LLM backbones with similar parameter sizes (from 7B to 13B), although there are also some "micro" models, such as VILA-1.5-3B-video, mini-Monkey, and Video-CCAM-4B, whose LLM backbones' parameter sizes are only 3B and 4B; additionally, we include image LLMs supporting **multi-image inputs**, such as VILA-1.5-13B (Lin et al., 2024), and those supporting **only single-image inputs**, like Monkey (Li et al., 2024b), mini-Monkey (Huang et al., 2024), and GLM-4V (GLM et al., 2024).

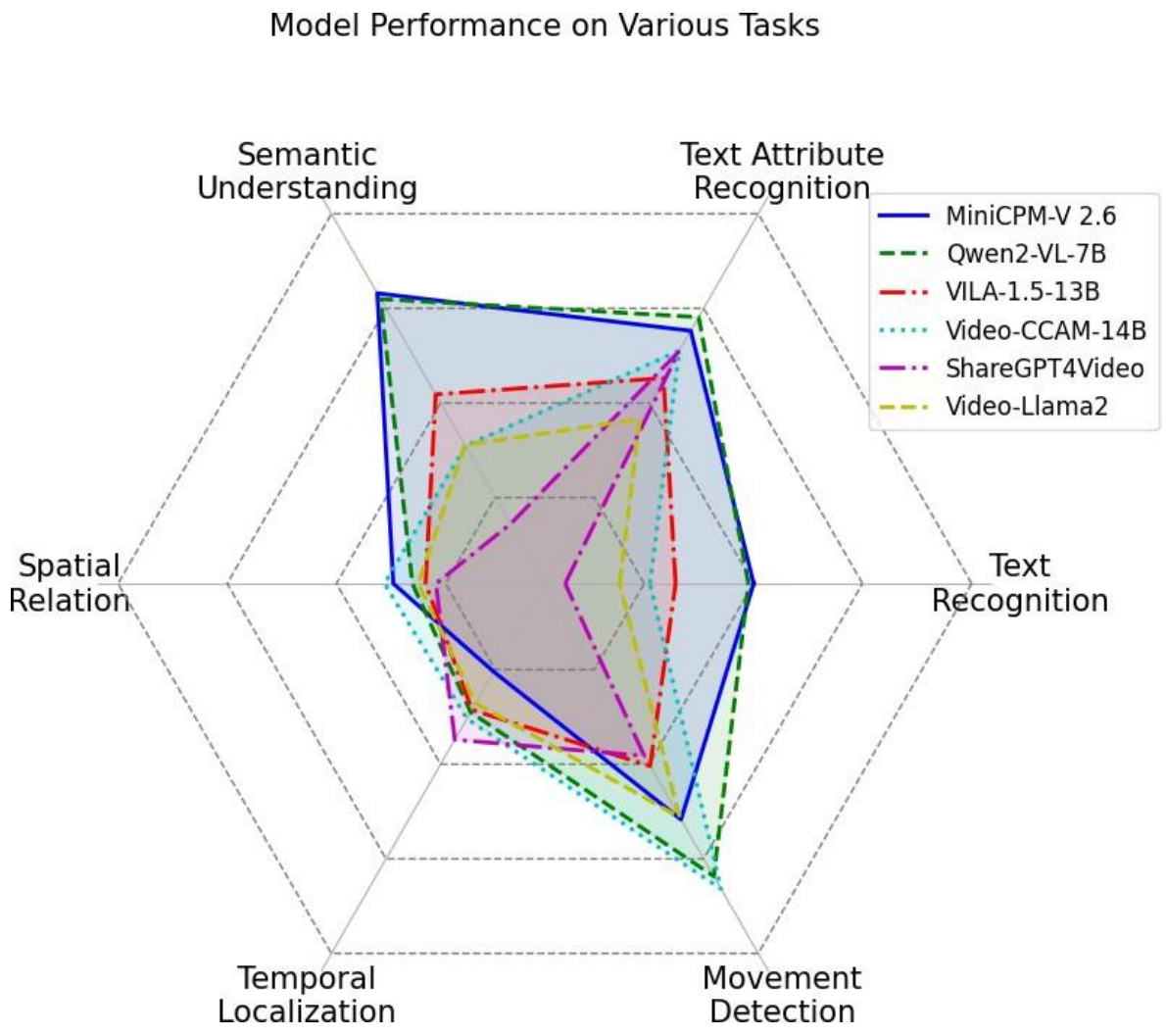

Figure 2: Overall performance of the six selected models on different sub-tasks in this benchmark.

#### 5.1.2 Metrics for diverse sub-tasks

| Sub-task's Type | Ground Truths' Diversity | Metric |
|---|---|---|
| Semantic Understanding | High | GPT-4o-mini |
| Spatial Relation | Moderate | GPT-4o-mini |
| Text Attribute Recognition | Low | Accuracy (Simple Check) |
| Text Recognition | Low | Accuracy (Simple Check) |
| Movement Detection | High | Accuracy (Multiple Choices) |
| Temporal Localization | Low | Accuracy (Multiple Choices) |

Table 1: Different Evaluation Metrics for 6 sub-tasks

Different sub-tasks need different metrics for model evaluation. For Text Recognition and Text Attribute Recognition, we selected accuracy, a simple way of checking the whether the ground truth is in the model's answer, as the evaluation metric.

Adopted by OCRbench (Liu et al., 2023b), This metric is particularly intuitive and concise.

For Semantic Understanding and Spatial Relation, different ways of expressions, variations in wording, and varying descriptions of objects or spatial-temporal locations can all lead to vastly different ground truth answers. So **GPT-4o-mini** is selected to calculate accuracy when dealing with synonymous expression situation.

In the Temporal Localization and Movement Detection sub-tasks, to lower task difficulty, same as in Video-MME (Fu et al., 2024), accuracy is simply calculated by the proportion of questions whose correct options are selected by models.

### 5.1.3 Ensure models output in proper format

To enable the model to better answer questions on some sub-tasks and ensure that the answers conform to the expected format or direction, we first added some appropriate prompts to the model along with the input questions. For example, in Text Recognition, Text Attribute Recognition, we require those models to output their answer in concise manner. In Movement Detection and Temporal Localization task, we design a simple prompt to require models answer questions only in the format of a single letter as the predicted option.

To evaluate LLMs which only support single image input, we simply concatenate sampled frames into a wider image horizontally, allowing it to represent the video as the model's input. It turns out that the three single-image-input models can tackle this kind of wide image and output normally at most time.

### 5.1.4 Overall Evaluation Results

In this evaluation process, we follow their official configurations and try to use more frames for evaluation. The numbers of the evenly sampled frames are 1 frame per second (up to 100 frames) for Chat-Univi-V1.5, 1 frame per second (up to 64) for MiniCPMV-v2.6, 1 frame per second (up to 768) for Qwen2-VL, 32 for ShareGPT4Video, 32 for Video-LLaMA2 (both 7B and 72B), 19 for VILA-1.5 (both 3B and 13B), 0.5 frame per second for Mini-GPT4video, 32 for Video-CCAM (both 4B and 14B), 32 for GLM4-V, 32 for Monkey and mini-Monkey.

On this benchmark, Qwen2-VL-7B performs best overall with the highest average score (0.4653), particularly excelling in text recognition, semantic understanding, text attribute recognition, and movement detection. MiniCPM-V 2.6 follows closely with an average score of 0.4508, with a highly small gap between the two, especially strong in semantic understanding and text recognition. While Qwen2-VL-7B has a slight edge in some tasks, both models show similar overall capabilities in handling video OCR tasks. In contrast, models like MiniGPT4-video, GLM-4V, mini-Monkey and Monkey perform significantly worse, with 3 of the 4 are single-image input LLMs, indicating clear limitations of this kind of visual LLMs in handling complex video OCR tasks.

### 5.1.5 Evaluation on different time interval thresholds

In the last table 2, we observed that all models do not perform well in Temporal Localization sub-task, with few model get accuracy significantly bigger than 0.25 (the accuracy of random guess). And it is weird that ShareGPT4Video perform best in this sub-task while do not have comparable ability in other sub-tasks. To further explore the capability of various nowadays' visual LLMs in timestamp localization, previous multi-choice format of data is abandoned which means models need to answer in the number of second to directly predict the timestamp. We select set 12 thresholds for judging whether models' prediction is within the pre-set margin of error (if so, a prediction will be seen as correct). The total number of question-answer pairs in this sub-task is 586.

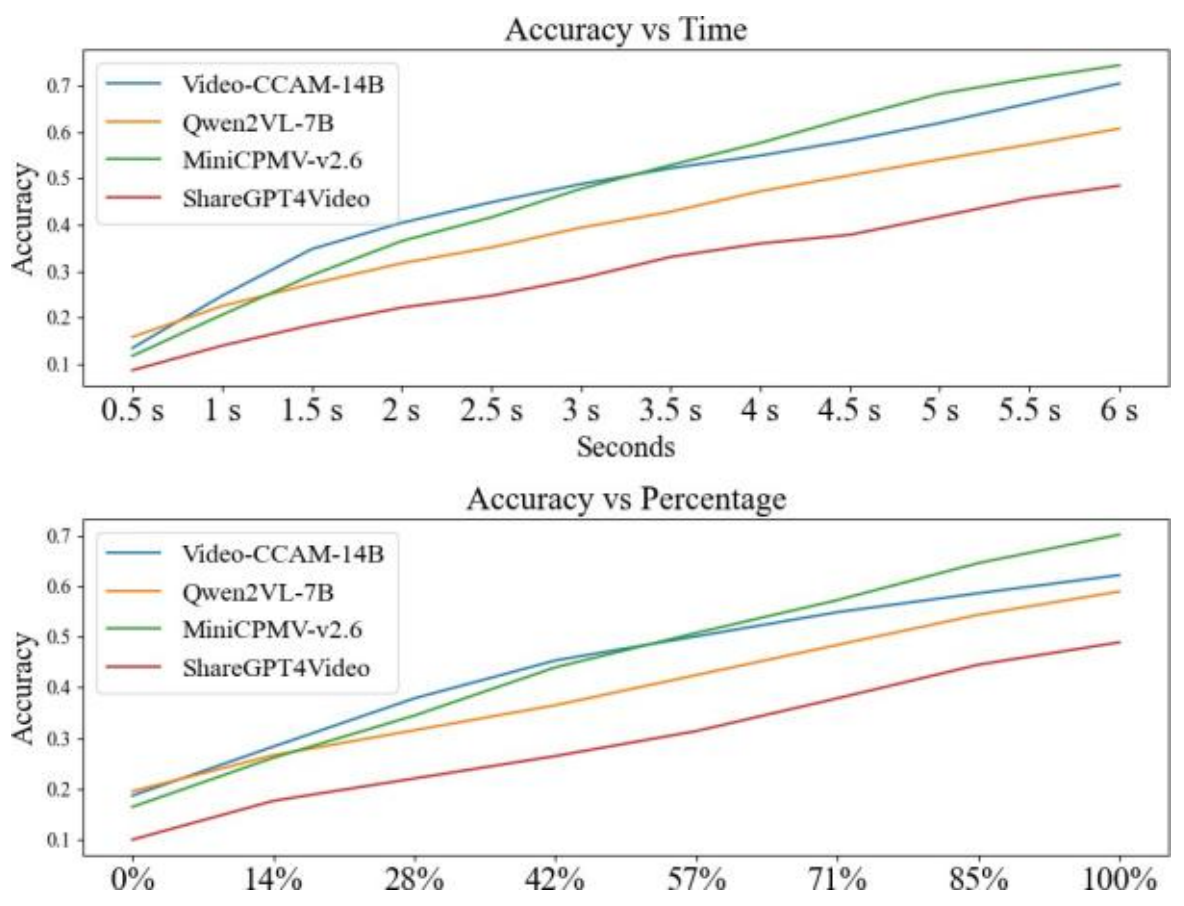

Figure 3: Top figure: change in accuracy with respect to the threshold (in seconds). Bottom figure: change in accuracy with respect to the threshold (as a percentage of the total video duration). 568 questions in Total.

We select 4 models (Qwen2-VL-7B, MiniCPM-V-2.6, Video-CCAM-14B, which have high average accuracy, and ShareGPT4Video that perform the best on multiple-choice Temporal Localization task) to evaluate. And we plot 2 line graph to show how the models' ability to locate timestamps changes with the threshold.

| Video OCR Tasks Leaderboard | | | | | | | | |
|---|---|---|---|---|---|---|---|---|
| **Models** | **LLM Backbone** | **TR** | **SU** | **SR** | **TAR** | **MD** | **TL** | **Average** |
| **Chat-UniVi-V1.5** | Vicuna-v1.5-7B | 10.27 | 23.77 | 17.20 | 28.68 | 11.30 | 13.82 | 17.51 |
| **MiniCPM-V 2.6** | Qwen2-7B | **40.07** | **63.19** | <u>29.57</u> | <u>55.26</u> | 51.68 | 30.72 | <u>45.08</u> |
| **VILA-1.5-3B-video** | LLaVA-VL | 21.47 | 34.92 | 18.01 | 35.26 | 35.82 | 24.91 | 28.40 |
| **Qwen2-VL-7B** | Qwen2-7B | <u>39.10</u> | <u>61.96</u> | 26.08 | **58.16** | 63.70 | 30.20 | **46.53** |
| **Video-LLaMA2-7B** | LLaMA2-7B | 15.50 | 31.09 | 25.00 | 36.58 | 51.20 | 27.13 | 31.08 |
| **Video-LLaMA2-72B** | LLaMA2-72B | 26.48 | 43.55 | 28.76 | 45.52 | **70.91** | **36.00** | 41.88 |
| **Mini-GPT-4-video** | LLaMA2-7B | 5.59 | 6.44 | 6.45 | 20.26 | 5.29 | 7.68 | 8.62 |
| **Video-CCAM-4B** | Phi-3-mini-4B | 17.66 | 23.58 | 25.00 | 38.95 | 20.19 | 19.80 | 24.19 |
| **Video-CCAM-14B** | Phi-3-medium-14B | 21.08 | 30.93 | **31.18** | 51.05 | <u>66.35</u> | 30.38 | 38.50 |
| **ShareGPT4Video** | LLaVA-Next-8B | 5.53 | 14.42 | 21.77 | 50.79 | 38.22 | <u>34.81</u> | 27.59 |
| **Monkey** | Qwen-7B | 3.74 | 6.29 | 8.06 | 25.26 | 32.69 | 29.52 | 17.61 |
| **Mini-Monkey** | InternLM2-1.8B | 13.42 | 23.93 | 16.40 | 46.05 | 42.07 | 28.50 | 28.41 |
| **GLM-4V** | GLM4-9B | 4.46 | 8.90 | 15.32 | 29.21 | 50.00 | 25.43 | 22.22 |
| **VILA-1.5-13B** | Vicuna-v1-13B | 25.76 | 41.81 | 2366 | 45.26 | 40.38 | 28.50 | 34.23 |

Table 2: Evaluation Results of Visual LLMs on Video-OCR Bench: The top-performing models are indicated in bold and the second-best models are marked with underlining. Values of accuracy are scaled up by a factor of 100 for better visualization.

From the figure it seems that although ShareGPT4Video performs the best on Temporal Localization sub-task when ground truths are time intervals being put in multiple options, it doesn't mean that this model really have the ability to perceive time due to its worst direct localization performance among the 4 chosen models. Besides, when the threshold is set as 2 seconds, same as length of options' time interval, the discrepancy of accuracy between these 2 setting is quite large.

#### 5.1.6 Re-Evaluation on single-image input LLMs

Our evaluation of single-image-input LLMs reveals relative poor performance, indicating that simple concatenation is not suitable for leveraging the models' OCR capabilities. To address this, we propose an alternative approach: instead of merging frames into a single visual input, we evaluate each frame independently frame by frame. We manually selected 506 question-answer pairs from the dataset, focusing on around 300 questions related to Text Recognition and several dozen each for Text Attribute Recognition, Semantic Understanding, and Spatial Relations. This selection was based on whether questions could be assessed frame-by-frame. Questions related to Movement Detection and Temporal Localization, which involve information spread across multiple frames, do not fit this approach. Similarly, questions involving temporal features or events were excluded.

To infer videos using single-image input LLMs, we propose a streamlined workflow. We first sample 10 evenly spaced frames from each video. Each frame, along with its corresponding question and prompt, is sequentially input into the model. If the model's response contains the ground truth answer, we mark the question as correctly answered and cease further frame testing. If the ground truth answer is not present, we continue testing the remaining frames and use GPT-4o-mini for final evaluation. Accuracy is computed based on the number of correctly answered questions.

We have collected the inference time and accuracy of the 3 models under this two different methods. The table show that processing frames individually generally leads to much higher average accuracy and significantly longer inference time compared to processing a single concatenated frame. It is noticeable that improvement in accuracy is quite large. This is mainly because some frames provide clear and easily recognizable view of OCR objects while others not. These kinds of view sometimes do not get enough attention when blending with other frames' visual features.

We conducted another experiment with the two

| Models | Setting | Inference Time | Accuracy |
|---|---|---|---|
| Monkey | ① | 4781 s | 0.502 |
| | ② | 1477 s | 0.065 |
| mini-Monkey | ① | 4176 s | 0.551 |
| | ② | 1550 | 0.219 |
| GLM-4V | ① | 8278 s | 0.592 |
| | ② | 1787 s | 0.115 |
| Qwen2VL | ③ | 2296 s | 0.603 |
| | ④ | 1852 s | 0.686 |
| | ① | 4127 s | 0.704 |
| MiniCPMV-v2.6 | ③ | 763 s | 0.589 |
| | ④ | 1283 s | 0.599 |
| | ① | 4830 s | 0.708 |

Table 3: Evaluation Result Comparison, ① represents evaluation frame by frame. ② represents evaluation use 1 concatenated frames. ③ represents evaluation with direct video input. ④ represents evaluation with sampled frames as multi-image input

strongest models, MiniCPMV-v2.6 and Qwen2VL-7B, which are the only models evaluated that support both direct video input and multi-image input. We assessed these models under three conditions: direct video input, sampled frames as multi-image input, and single frames processed sequentially.

Our results, as shown in Table 3, reveal that performance varies significantly with different evaluation settings. Both Qwen2VL and MiniCPMV-v2.6 demonstrate a marked improvement in accuracy when evaluated frame-by-frame compared to direct video input and multi-image input. This indicates that processing videos frame-by-frame enhances the models' ability to capture detailed visual features, leading to more accurate predictions. The improvement may be attributed to the challenge of isolating targeted visual features when multiple images are combined.

## 6 Conclusion

In this paper, we introduce **Video-OCR bench**, the first comprehensive benchmark to evaluate visual LLMs in various video-OCR tasks. Our benchmark presents new tasks, ranging from text-related movement tracking to time stamp localization. We also propose a GPT-4o-mini model-based evaluation method, inspired by Video-ChatGPT (Maaz et al., 2023), for assessing some complicated tasks. Our evaluation result shows that (1) many video LLMs struggle with video OCR tasks. Image LLMs, although show good capacity in several sub-tasks like Text Recognition and Semantic Understanding, their ability to handle video content is relatively weak. (2) All models perform poorly in Temporal Localization, with accuracy not better than random guess. These findings underscore the need for further advancements in handling features about temporal information and motion tracking in dataset. We hope this Video-OCR bench will inspire future research and development in improving the capabilities of visual LLMs.

## 7 Future Work

Most evaluated visual LLMs exhibit inadequate performance in timestamp localization, suggesting that while they can capture temporal information about objects and events, their understanding of time itself is limited. Additionally, many models struggle with motion tracking. To address these issues, it is crucial to expand the benchmark to include the full dataset, including training data, and explore fine-tuning of video LLMs. This process should also involve evaluating whether current visual LLMs architectures can be effectively trained for these tasks or if a new model architecture is needed to better capture relevant features.

## Limitations

In this study, we employed a method for processing video inputs using single-image-input LLMs, where instead of summarizing predictions across multiple frames—commonly achieved by leveraging tools like GPT-4-omini—we utilized the ground truth to bypass the summarization step. While this approach simplifies the workflow, it introduces two significant limitations. First, the inference time is likely underestimated since not only the time required for summarization is omitted from the pipeline but also not all frames are inferred by models. Second, the model's accuracy may be a little higher. Furthermore, due to both financial and computational resource constraints, we only perform 1 model with larger parameter scales, i.e. Video-LLaMA2-72B. This may limit the generalizability of our findings across larger and more complex architectures. Future work will address these limitations by incorporating more comprehensive evaluation strategies and extending experiments to include larger models.

## Acknowledgments

This research was supported by *National Natural Science Foundation of China* (No.62406121) and *Natural Science Foundation of Hubei Province, China* (No.2024AFB189).

# A Data Statistics

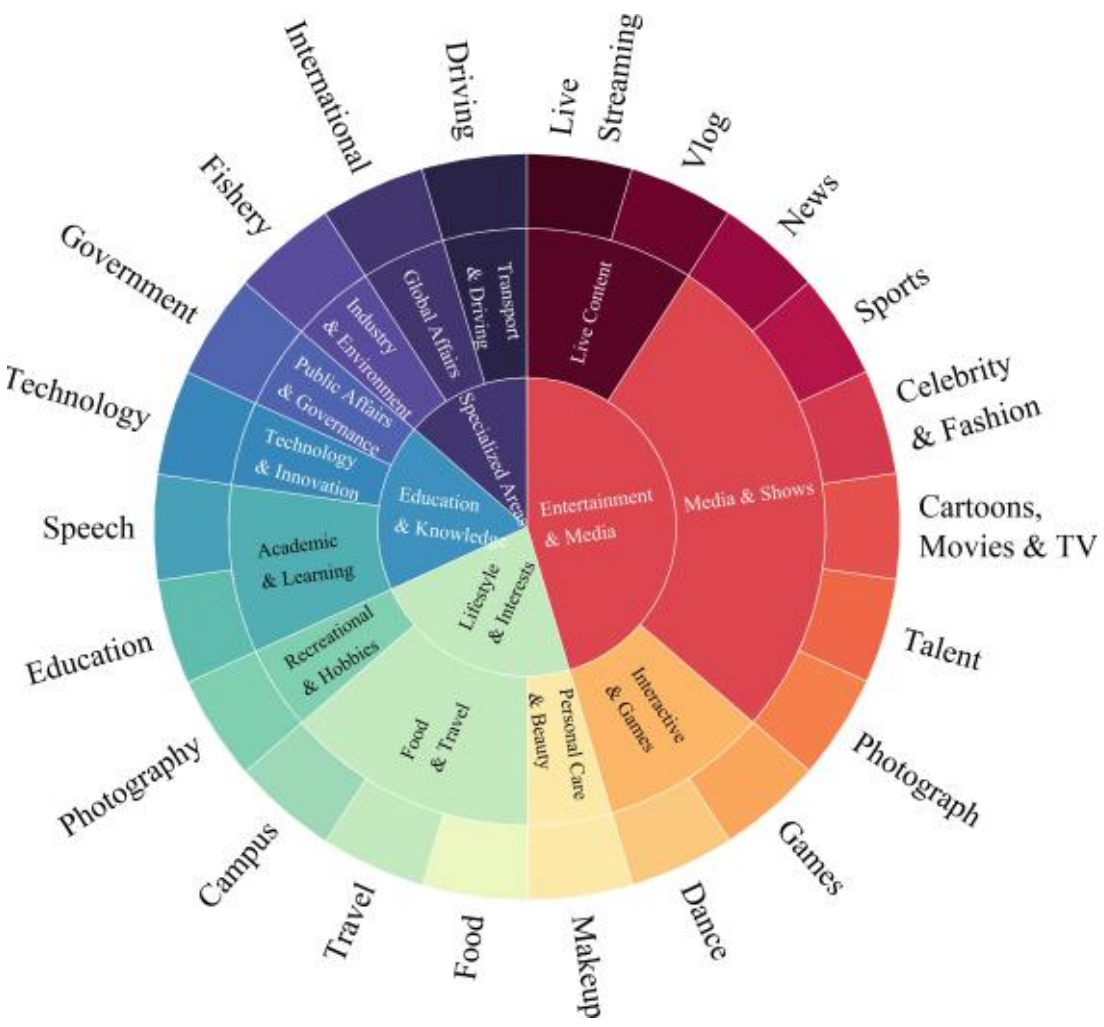

Figure 4: Distribution for question-answer pairs in various sub-tasks. Color labels represent words count in questions.

Here, we present the detailed statistics of our dataset to provide a more comprehensive understanding, including the meta information, QA pairs, video clips, qualitative analysis, and comparison to previous works.

### A.1 Video Distribution

Here is a pie chart 4 the distribution of duration and scene of videos in dataset. To ensure the diversity of video scenes, we obtained metadata from 4 different source datasets. As a result, our video scene range covers videos in more than 20 different scenarios, including: Livestreaming, Sports, Celebrity & Fashion, Game, Cartoon & Movie & TV show, Interview, Introduction, Photograph, Speech, News, Driving, Vlog, and so on.

Across this videos from diverse types of scenario, videos shot during driving occupied the most. This kind of video provide much good enough data due to texts' enough motion. This is important because it provides a dynamic perspective for stationary text and related objects. The disappearance of text will be gradual rather than sudden. When the scene text is rich and contains sufficient semantic information, relative movement can also cause issues such as blurriness, improper exposure, and occlusion. These issues, along with blurry and incomplete text, will pose challenges to the model's video OCR capabilities in terms of robustness.

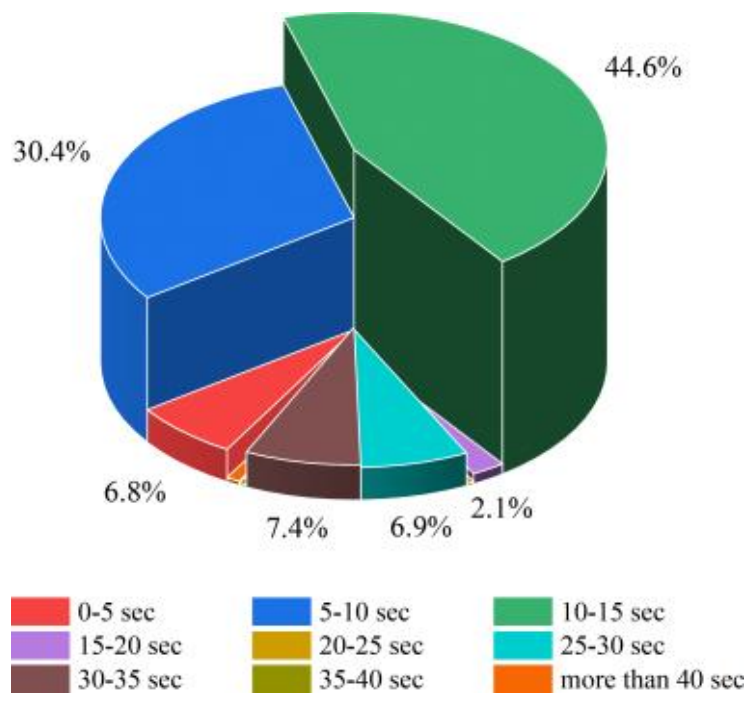

Figure 5: Distribution for question-answer pairs in various sub-tasks. Color labels represent words count in questions.

As for the duration of selected video, our selected videos mainly last from a little over ten seconds to several tens of seconds.

### A.2 QA pairs distribution

Above is distribution of question-answer pairs related to videos in benchmark. Due to the fact that text recognition task is fundamental and semantic understanding task is more important than others, we collect much more data than other sub-tasks. In this figure 7 we can observe that most questions in this benchmark have less than 15 words, which is intentionally designed for contain enough information for precise locating OCR objects with least words.

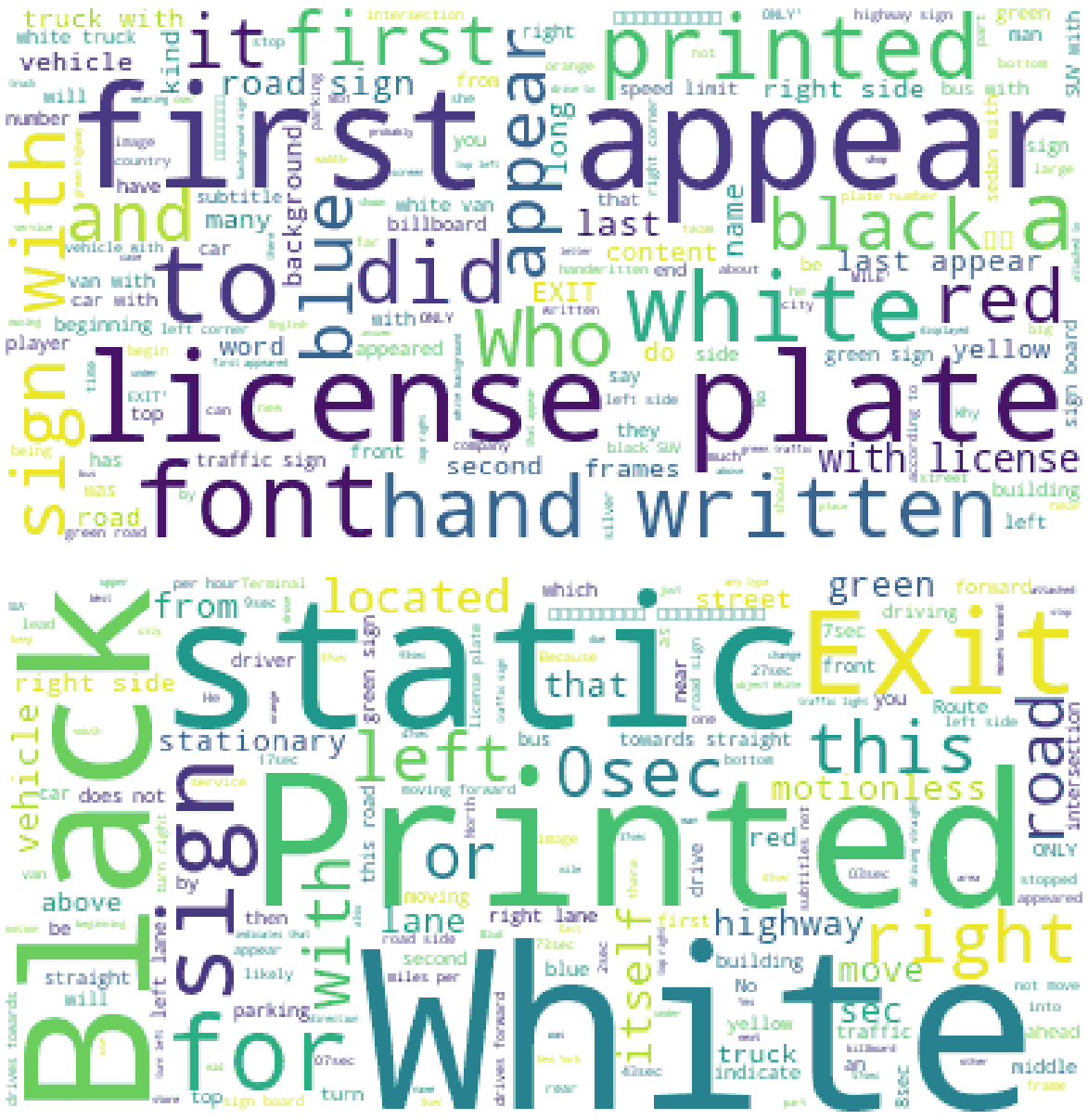

Figure 6: Questions Word cloud (top) and Answers Word cloud (bottom).

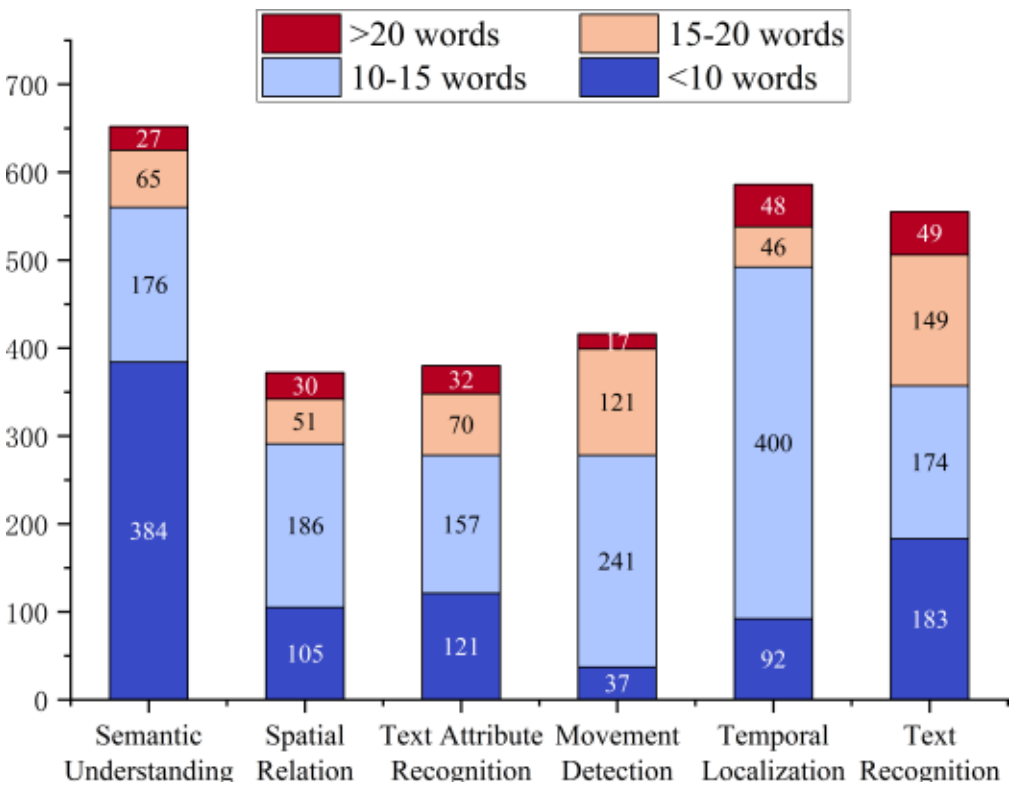

Figure 7: Distribution for question-answer pairs in various sub-tasks. Color labels represent words count in questions.

# B Format of traditional video OCR data and Diagram for Semi-Automatic Process

## B.1 traditional video OCR data format

This JSON data structure is commonly used in traditional video OCR systems to store information extracted from video frames. It captures the video name as a unique identifier, the file path of the image frame, and an array of OCR labels. Each OCR label provides details about the recognized text blocks, including their ID, content, attributes (such as language), and bounding box coordinates that define their location within the image. See figure 8.

```json
{
    "video_name": 701,
    "frame_path": "/701/frames_0001.jpg",
    "ocr_labels": [
        {
            "text_id": 303,
            "text_content": "T683270C",
            "Attribute": "English_Text",
            "bounding_box": {
                "x1": 389.9609,
                "x2": 475.3973,
                "y1": 442.9447,
                "y2": 472.0858
            }
        },
        ......
    ]
}
```

json

Figure 8: traditional video OCR data format

## B.2 Diagram of Semi-Automatic

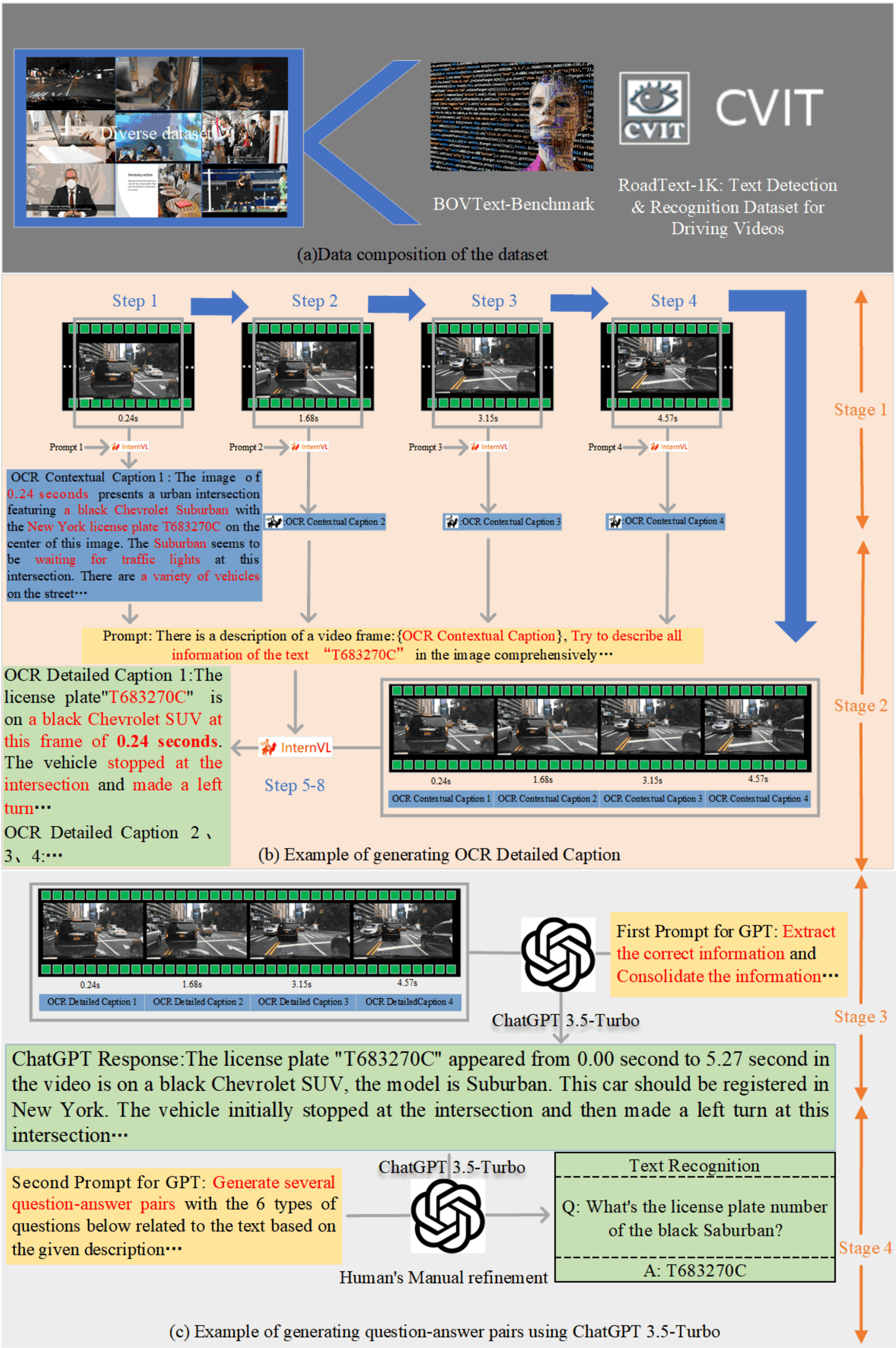

Figure 9: Diagram for semi-automatic process of data manufacture

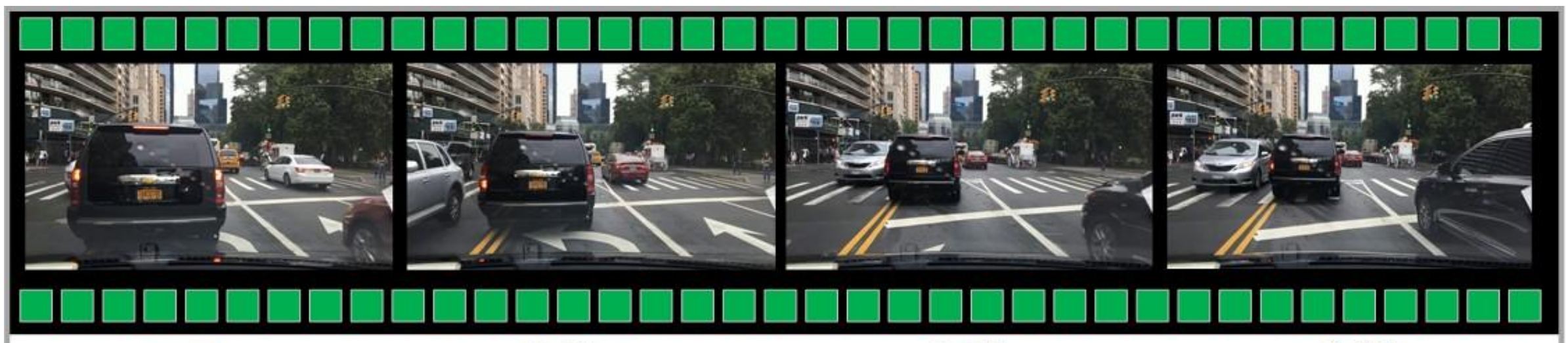

0.24s 1.68s 3.15s 4.57s

**Prompt for Frame 1(Input InternVL):**

Describe all the visual information in a given image comprehensively. The image you receive is a video frame sampled from a video. Besides, try to include information of these text in the image as well.
(1): "T683270C", appeared at the center of the image.
(2): "SUBURBAN", appeared near the mid-left area of the image.
(3): "Park", appeared near the top-left corner of the image.
(4): "PHOENIX", appeared near the top-left area of the image.
(5): Another "PHOENIX", appeared near the top-left area of the image.

**OCR Contextual Caption For Frame 1(InternVL Output):**

The image of 0.24 seconds presents a bustling urban intersection featuring a black Chevrolet Suburban with the New York license plate T683270C on the center of this image. The Suburban seems to be waiting for traffic lights at this intersection. There are a variety of vehicles on the street, including a white sedan, a yellow taxi, and a white carriage on the far right. To the left, two signs reading "PHOENIX" and "park" suggest nearby commercial establishments. Above, traffic lights show green, guiding the steady flow of vehicles past towering skyscrapers that define the city's skyline. Pedestrians traverse the crosswalks, contributing to the lively atmosphere of this dynamic urban environment.

**Prompt for Frame 1(Input InternVL):**

You will be given a video frame sampled from a video, shot from driver's perspective on American roads. There is a description of a video frame: {**OCR Contextual Caption For Frame 1**}. Based on this description and given image, Try to describe all information of the text "T683270C" in the image comprehensively. You may find it at center of the frame.
ATTENTION:
Your answer should include as many details as possible, such as:
(1) the font color and background color of the text, and whether the text is hand-written or printed,
(2) the object on which the text is located, and how this object moves in the video,
(3) this text's semantic meaning in the image.

**OCR Detailed Caption 1 for Frame 1(InternVL Output):**

The license plate "T683270C" appeared from 0.00 second to 5.27 second in the video is on a black Chevrolet SUV. the model is Suburban. This car should be registered in New york. The vehicle initially stopped at the intersection and then made a left turn at this intersection as shown by its shining left-turn light. There are two parking signs close to the intersection in the upper left corner of the video frame. The intersection is bustling with a lot of traffic, and in the distance, there is a white carriage, This intersection is likely to be somewhere in the area of New york City.

Figure 10: Example of 2 kinds of Prompts to generate OCR Contextual Caption and OCR Detailed Caption respectively.

**First Prompt for GPT(Input GPT) :**

You will receive 4 descriptions of the same text appeared in a video. Each description follows a similar format and includes information about the text's font color, the background color on which the text is placed, the object on which the text is located, how the object moves, etc. These descriptions may contain discrepancies in their details.
Your task is to:
1. Extract the correct information based on the majority rule: If a majority of the descriptions agree on a particular detail, select that detail as the correct one. If there is no majority, use your best judgment to determine the most likely correct information. If one description has a new information which all other descriptions do not contain, treat it as illusion and ignore it.\n"
2. Consolidate the information:
Combine the correct details into a single, coherent description.\n"
Do not begin your answers with any preamble or prefixes.\n"
Here is an example of how you should respond to user's request:\n"
"------\n"
##Case 1:\n"
User: 4 descriptions are below:\n"
The text is blue, placed on a white background, and is a license plate of a black SUV, specificly Chevrolet Saburban.\n
The text is black, placed on a white background, and is a license plate located on a black Subaru SUV Saburban.\n
The text is black, placed on a grey background, and is a license plate located on a black Chevrolet Saburban
The text is green, placed on a white background, and located on a plastic sign.\n
Assistant: The text is black, placed on a white background, and is a license plate located on a black Chevrolet SUV Saburban.

**Second Prompt for GPT(Input GPT):**

You will be given a description of a certain text appearing in a video. Your task is to generate several question-answer pairs with the 6 types of questions below related to the text based on the given description.
The 6 types of questions you should generate include:
1. Text Recognition: Questions about the text itself, like 'What's the text on the road sign?'.
2. Semantic Understading: Questions about the Semantic Role and Contextual meaning of the text, like 'What does the text {given_text} located on the road sign mean in the video?'.
3. Spatial Relation: Questions about the text's spatial relationship with other objects in video, like 'What's the object where the text {given_text} is located?', 'What is kind of the shop under the text {given_text}?'.
4. Text Attribute Recognition: Questions about the font color and background color of the text, such as 'whether the text {given_text} is hand-written or printed', and 'What's the font color of {given_text}?'.
5. Movement Detection: Questions about how the text changes its spatial position in the video frame, or how the object with the text moves, like 'How does the text {given_text} move in video frames?', 'How doesthe object with the text moves'.
6. Temporal Localization: Questions about the temporal information of the text's appearance in the video, like 'How long does the text {given_text} last in the video?', 'When does the text {given_text} first appear and disappear in the video?', and 'When does the text {given_text}last appear in the video?'.
ATTENTION:
1. The types of information should be appended to the answers.
2. The nouns, things, and objects mentioned in all questions should be clearly identified due to potential similar objects in a video.
3. Do not begin your answer with any preamble, prefixes, transitional sentences, or lead-in sentences such as "Here is the question-answer pair:".
4. Give your answer in JSON format directly! Each question and answer pair should be formatted as follows:

```
{
    "1":{
        "Question Type": "Type of question",
        "Q": "The question",
        "A": "The answer"
    },......
}
```

5. Ensure the entire output is a valid JSON string.

Figure 11: The First is an example of prompts to aggregate OCR Detailed Captions. The Second is an example of prompts to generate question-answer pairs for 6 sub-tasks.

Here are 2 cases of the format how you should respond to the user's request:

Case 1:

User: The text "32 605 94" located at the center of the video frame is a License Plate in black font with a yellow background. It appears in the video at 0.00sec - 7.87sec. This license plate is likely on the rear of a black Toyota SUV Land Cruiser.

Assistant:

{

"1": {"Question Type": "Text Recognition", "Q": "What is the license plate of the Toyota SUV in the center of the video frame?", "A": "32 605 94."},

"2": {"Question Type": "Semantic Understanding", "Q": "What is the meaning of the text 32 605 94 in the center of the video frame?", "A": "It is a License Plate number for vehicle registration."},

"3": {"Question Type": "Text Attribute Recognition", "Q": "What is the font color of the License plate 32 605 94?", "A": "Black"},

"4": {"Question Type": "Spatial Relation", "Q": "What kind of vehicle is the license plate 32 605 94 located on?", "A": "A black Toyota SUV Land Cruiser."},

"5": {"Question Type": "Movement Detection", "Q": "How does the license plate 32 605 94 move in the video frame?", "A": "It stays at the center of the video frame."},

"6": {"Question Type": "Temporal Localization", "Q": "When does the license plate 32 605 94 first appear in the video?", "A": "0.52sec."}

}

Case 2:

User: The text "NYPD" located at the mid-right of the video frame is a blue text on the left-gate side of a white van. It appears in the video at 0.00sec - 4.89sec. It's an abbreviation of "New York Police Department", indicating that the van is a New York police vehicle. The white van seems turning left.

Assistant:

{

"1": {"Question Type": "Text Recognition", "Q": "What is the text on the left-gate side of the white van in the video frame?", "A": "NYPD."},

"2": {"Question Type": "Semantic Understanding", "Q": "What is the meaning of the text NYPD on the left-gate side of the white van in the video frame?", "A": "It is an abbreviation of New York Police Department."},

"3": {"Question Type": "Spatial Relation", "Q": "Which side of the white van is the text NYPD located on?", "A": "The left."},

"4": {"Question Type": "Text Attribute Recognition", "Q": "What is the font and background color of the text NYPD?", "A": "Blue and White."},

"5": {"Question Type": "Spatial Relation", "Q": "What kind of vehicle is the text NYPD located on?", "A": "A white van."}

}

Figure 12: Few shot cases for generating question-answer pairs via GPT-3.5